\pdfoutput=1

\documentclass[11pt]{article}

\usepackage{acl}

\usepackage{times}
\usepackage{latexsym}

\usepackage{amsmath,amsfonts,bm,amssymb}
\usepackage{booktabs} %
\usepackage{url}
\usepackage{graphicx}
\usepackage{array}
\usepackage{color}
\usepackage{makecell}
\usepackage{multirow}
\usepackage{xspace}
\usepackage{colortbl}

\newcommand{\ie}{\textit{i.e.}} %
\newcommand{\eg}{\textit{e.g.}} %
\newcommand{\start}[1]{\vspace{.0mm}\noindent{{\bf #1}.}}

\newcommand{\upv}{\vspace{-.0cm}}
\newcommand{\downv}{\vspace{-.0cm}}

\newcommand{\ours}{\textsc{UniPELT}\xspace}
\newcommand{\oursAP}{\textsc{UniPELT} (AP)\xspace}
\newcommand{\oursAPL}{\textsc{UniPELT} (APL)\xspace}
\newcommand{\task}{PELT\xspace}

\definecolor{gred}{RGB}{219,68,55}
\definecolor{gblue}{RGB}{76, 159, 229}
\definecolor{gyellow}{RGB}{244,180,0}
\definecolor{ggreen}{RGB}{85,157,88}
\definecolor{ggrey}{RGB}{115,115,115}
\definecolor{na}{gray}{0.9}

\newcommand{\colorB}[1]{\textcolor{gblue}{\textbf{#1}}}

\usepackage{amsmath,amsfonts,bm}

\def\eqref#1{equation~\ref{#1}}

\def\1{\bm{1}}

\def\vh{{\bm{h}}}

\def\mK{{\bm{K}}}

\def\mP{{\bm{P}}}
\def\mQ{{\bm{Q}}}

\def\mV{{\bm{V}}}
\def\mW{{\bm{W}}}

\DeclareMathAlphabet{\mathsfit}{\encodingdefault}{\sfdefault}{m}{sl}
\SetMathAlphabet{\mathsfit}{bold}{\encodingdefault}{\sfdefault}{bx}{n}

\def\gG{{\mathcal{G}}}

\def\gM{{\mathcal{M}}}

\def\sR{{\mathbb{R}}}

\usepackage[T1]{fontenc}

\usepackage[utf8]{inputenc}

\usepackage{microtype}

\title{\ours: A Unified Framework for Parameter-Efficient \\Language Model Tuning}

\author{\makecell{Yuning Mao$^{1}$\thanks{\hspace{.06in}Work was done during internship at Meta AI.}, Lambert Mathias$^{2}$, Rui Hou$^{2}$, Amjad Almahairi$^2$,\\ Hao Ma$^2$, Jiawei Han$^1$, Wen-tau Yih$^2$, Madian Khabsa$^2$} \\

$^1$University of Illinois Urbana-Champaign  \quad \{yuningm2, hanj\}@illinois.edu \\ 
$^2$Meta AI  \quad \{mathiasl, rayhou, aalmah, haom, scottyih, mkhabsa\}@fb.com
}

\begin{document}
\maketitle
\begin{abstract}
Recent parameter-efficient language model tuning (\task) methods manage to match the performance of fine-tuning with much fewer trainable parameters and perform especially well when training data is limited.
However, different \task methods may perform rather differently on the same task, making it nontrivial to select the most appropriate method for a specific task, especially considering the fast-growing number of new \task methods and tasks.
In light of model diversity and the difficulty of model selection, we propose a unified framework, \ours, which incorporates different \task methods as submodules and learns to activate the ones that best suit the current data or task setup via gating mechanism. 
On the GLUE benchmark, \ours consistently achieves 1\textasciitilde4\% gains compared to the best individual \task method that it incorporates and outperforms fine-tuning under different setups. 
Moreover, \ours generally surpasses the upper bound that takes the best performance of all its submodules used individually on each task, indicating that a mixture of multiple \task methods may be inherently more effective than single methods.\footnote{Our code can be found at \url{https://github.com/morningmoni/UniPELT}.}
\end{abstract}

\section{Introduction}
As pre-trained language models (PLMs) \cite{devlin-etal-2019-bert} grow larger and larger \cite{brown2020language}, it becomes increasingly infeasible to perform conventional fine-tuning, where separate replicas of the model parameters are modified per single task.
To solve the issue, there has recently been a surge of studies on \textbf{p}arameter-\textbf{e}fficient \textbf{l}anguage model \textbf{t}uning (\task), namely how to effectively tune the PLMs with fewer trainable parameters.

\begin{figure}[t]
\vspace{-.2cm}
    \centering
    \includegraphics[width=0.75\linewidth]{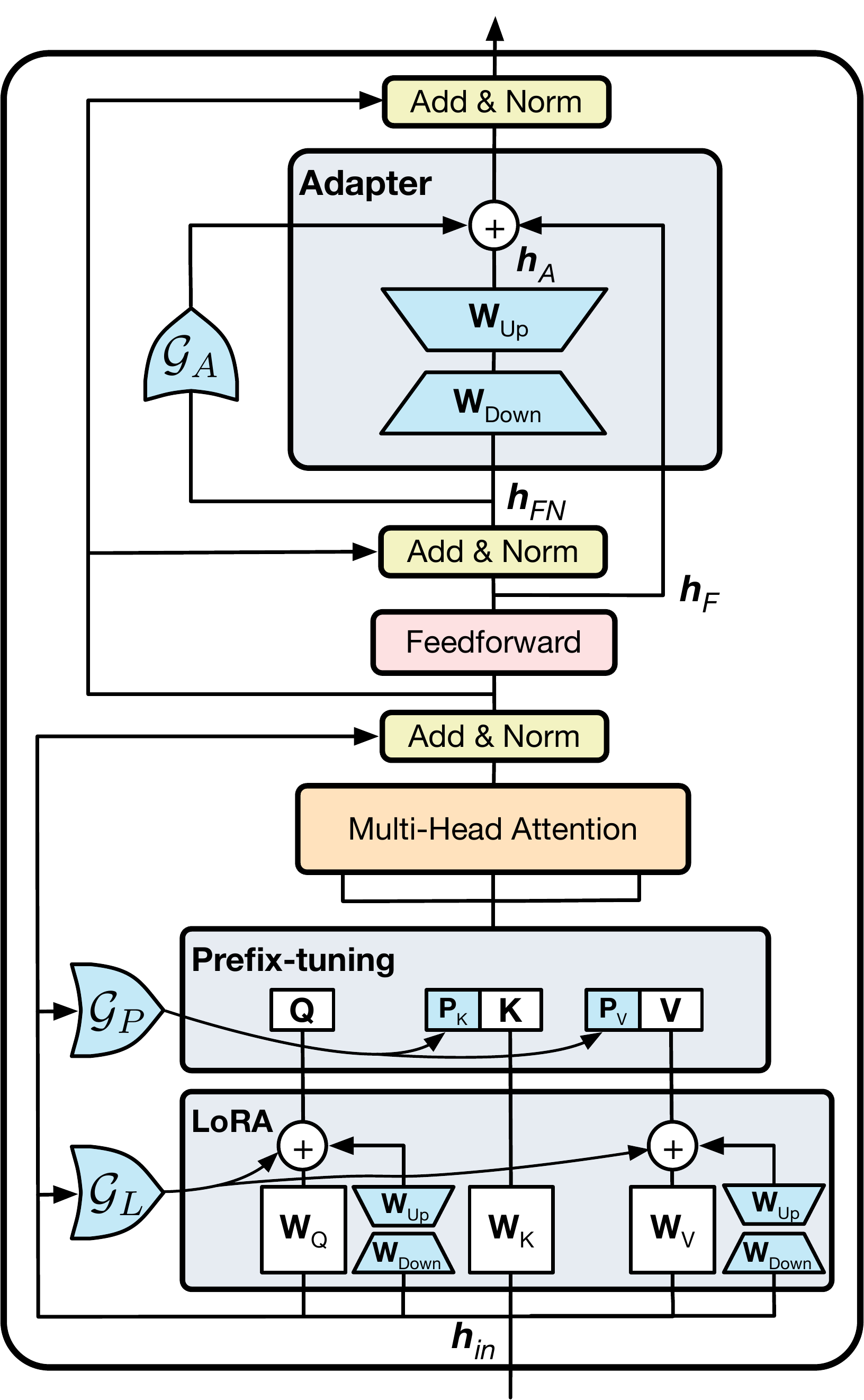}
    \upv
    \caption{Illustration of \ours, which subsumes existing PELT methods as submodules and controls them via gating mechanism $\gG$. Different (combinations of) submodules can be activated for different samples. The \colorB{trainable parameters} are shown in blue.}
    \label{fig_model}
    \downv
    \vspace{-.25cm}
\end{figure}

Existing \task research generally aims at achieving performance comparable to fine-tuning with as few trainable parameters as possible, which has seen significant progress -- the task-specific trainable parameters used in most recent approaches \cite{lester2021power,guo-etal-2021-parameter} are almost negligible compared to the total parameters of the PLM (<1\%).
A more challenging yet less studied problem is whether one can achieve better performance than fine-tuning with fewer parameters.
Recent studies \cite{he-etal-2021-effectiveness,li-liang-2021-prefix,karimi-mahabadi-etal-2021-parameter} find that some \task methods are more effective than fine-tuning on certain tasks when training data is limited, possibly due to the reduced risk of overfitting.
However, as found in our experiments (Table~\ref{table_glue_low_major}), different \task methods  exhibit diverse characteristics and perform rather differently on the same task, which makes it nontrivial to select the most appropriate method for a specific task, especially considering the fast-growing number of new \task methods and tasks \cite{pelt_github}.

In light of the diverse performance of \task methods and the cost of selecting the best method, we propose a unified \task framework, named \ours, which incorporates different \task methods as submodules and learns to dynamically activate the (combination of) submodules that best suit the current data or task setup.
As a result, model selection is no longer needed and consistently better performance is achieved under different setups. 
The activation of each submodule in \ours is controlled by \textit{gating mechanism}, which learns to favor (assign more weight to) the submodules that positively contribute to a given task. 
In addition, since the number of parameters introduced by each submodule is generally small, combining multiple methods leads to negligible losses in model efficiency.

We select four representative \task methods for our study -- adapter \cite{houlsby2019parameter}, prefix-tuning \cite{li-liang-2021-prefix}, LoRA \cite{hu2021lora}, and BitFit \cite{ben2021bitfit}, which largely cover the major categories of \task methods.
We perform two sets of analysis that carefully examines (i) the characteristics of individual \task methods and (ii) their effectiveness when coordinated by \ours under various setups.\footnote{BitFit is not included in \ours as it typically performs the worst in our preliminary experiments.}

Extensive experiments on the GLUE benchmark~\cite{wang2019glue}, with 32 setups (8 tasks $\times$4 data sizes) and 1,000+ runs, not only reveal the diverse behavior of \task methods, but also show that \ours is more effective and robust than using each method alone in various task and data setups.
Specifically, \ours consistently improves the best submodule that it incorporates by 1\textasciitilde4 points and even outperforms fine-tuning, achieving the best average performance on the GLUE benchmark under different setups.
Moreover, \ours generally surpasses the upper bound that takes the best performance of all its submodules used individually on each task, which suggests that \ours maintains (near) optimal performance under different setups.
The fact that \ours outperforms the upper bound also implies that a mixture of \task methods involving different parts of the PLM architecture may be inherently more effective than individual methods.

\start{Contributions}
(1) We conduct a comprehensive study of representative \task methods and carefully examine their differences and commonalities in terms of performance and characteristics.
(2) We propose a unified \task framework that can incorporate existing methods as submodules and automatically learn to activate the appropriate submodules for a given task.
(3) Our proposed framework achieves better average performance than fine-tuning and the \task methods that it incorporates under various setups, often performing the best and never the worst at per-task level, exhibiting superior effectiveness and robustness with negligible losses in model efficiency.

\section{Preliminaries}
\label{sec_prelim}

\subsection{\task Methods without Additional Parameters}
PLMs can be used as feature extractors where only the top layers or prediction head are fine-tuned without additional parameters \cite{lee2019would}. However, such fine-tuning approaches generally lead to degenerate model performance that is much worse than fine-tuning all parameters \cite{lee2019would,pfeiffer-etal-2021-adapterfusion}.
A recent method BitFit \cite{ben2021bitfit} only tunes the bias terms of the PLM and is shown to achieve performance comparable to fine-tuning on certain tasks when training data is limited.
Therefore, we select BitFit as the representative of this category for analysis.

\subsection{\task Methods with Additional Parameters}
Alternatively, one may fix the entire PLM and introduce a small number of new trainable parameters.
Notable examples in this category include adapter \cite{houlsby2019parameter} and its extensions \cite{pfeiffer-etal-2021-adapterfusion,karimi-mahabadi-etal-2021-parameter}, prefix-tuning \cite{li-liang-2021-prefix} and its extensions \cite{lester2021power}, and additive methods \cite{guo-etal-2021-parameter,hu2021lora}.

Next, we will briefly describe these methods to facilitate the introduction of our proposed framework.
An illustration is shown in Fig.~\ref{fig_model} for better understanding.

\start{Adapter}
Adapter \cite{houlsby2019parameter} adds a trainable \textit{bottleneck layer} after the feedforward network in each Transformer layer of the PLM.
A bottleneck layer consists of a down+up projection pair that shrinks and recovers the size of token hidden states.
Mathematically, if we denote the output of the \textbf{f}eedforward network after residual connection and layer \textbf{n}ormalization as $\vh_{FN}$ with hidden size $D_{\text{hidden}}$ and bottleneck size $D_{\text{mid}}$, then the output of a bottleneck layer $\vh_A$ is:
\begin{equation}
    \vh_A= \mW_{\text{up}}^\intercal \phi(\mW_{\text{down}}^\intercal \vh_{FN}),
\end{equation}
where $\mW_{\text{down}} \in \sR^{D_{\text{hidden}} \times D_{\text{mid}}}$, $\mW_{\text{up}} \in \sR^{D_{\text{mid}} \times D_{\text{hidden}}}$, $\phi$ is a nonlinear activation function, and the bias terms are omitted for brevity.
The parameters in layer normalization and the final prediction head sometimes are also fine-tuned depending on the specific adapter variants.

Adapter has shown to be on par with fine-tuning and sometimes exhibits better effectiveness in the low-resource setting \cite{he-etal-2021-effectiveness}.
Later studies extend adapter to multi-lingual \cite{pfeiffer-etal-2020-mad} and multi-task \cite{karimi-mahabadi-etal-2021-parameter} settings, or further reduce its trainable parameters \cite{mahabadi2021compacter}, which can be easily incorporated into \ours as a replacement of the vanilla adapter.

\start{Prefix-tuning}
Prefix-tuning \cite{li-liang-2021-prefix} prepends a number of task-specific trainable vectors to the input of multi-head attention in \textit{each} Transformer layer, which the original tokens can attend to as if they were virtual tokens.
Specifically, we denote the original sequence length $L_0$, the number of trainable vectors (\ie, prefix length) $L$, and the Transformer layer input $\vh_{\text{in}} \in \sR^{D_{\text{hidden}} \times L_0}$.
First, three linear projections $\mW_Q$, $\mW_K$, $\mW_V  \in \sR^{D_{\text{hidden}} \times D_{\text{hidden}}}$ transform $\vh_{\text{in}}$ into Query $\mQ$, Key $\mK$, and Value $\mV$.
Then, two prefix matrices $\mP_K$ and $\mP_V \in \sR^{D_{\text{hidden}} \times L}$ are prepended to $\mK$ and $\mV$.
To stabilize optimization, the prefix matrix $\mP$ is reparameterized by a feedforward network:
\begin{equation}
    \mP' = \mW_{\text{up}}^\intercal \phi(\mW_{\text{down}}^\intercal \mP),
\end{equation}
where $\mW_{\text{down}} \in \sR^{D_{\text{hidden}} \times D_{\text{mid}}}$, $\mW_{\text{up}} \in \sR^{D_{\text{mid}} \times 2 N_{\text{layer}} D_{\text{hidden}}}$, and $N_{\text{layer}}$ denotes the number of Transformer layers. 
The parameters of this network can be discarded after training, and only $2N_{\text{layer}}$ prefix matrices $\in \sR^{D_{\text{hidden}} \times L}$ are needed (2 matrices for each layer).

Prefix-tuning is originally evaluated on natural language generation and we adapt it to understanding tasks.
A follow-up method named prompt-tuning \cite{lester2021power} further reduces task-specific parameters by limiting the prefix to the first layer but only performs competitively with very large model sizes (billions of total parameters), and is thus not considered in our study.
Note that prefix-tuning (or prompt-tuning) is different from prompt-based \textit{fine-tuning} methods \cite{schick-schutze-2021-exploiting,gao-etal-2021-making} (see App.~\ref{sec_prefix_vs_prompt} for specific differences).

\start{Additive Methods}
Additive \task methods treat the model parameters after fine-tuning as an addition of the pre-trained parameters $\theta_{\text{pre-trained}}$ and task-specific differences $\delta_{\text{task}}$, where $\theta_{\text{pre-trained}}$ is fixed and a new (sub)set of model parameters are added on top: $\theta_{\text{task}} = \theta_{\text{pre-trained}} + \delta_{\text{task}}$. 
There are various ways to parameterize $\delta_{\text{task}}$, leading to different additive methods such as LoRA~\cite{hu2021lora}, diff pruning \cite{guo-etal-2021-parameter}, and side-tuning \cite{zhang2020side}.
We take LoRA as a representative and incorporate it into \ours. Other methods are conceptually similar and can be incorporated in the  same fashion.

LoRA introduces trainable low-rank matrices and combines them with the original matrices in the multi-head attention.
Specifically, two matrices $\mW_{\text{down}} \in \sR^{D_{\text{hidden}} \times D_{\text{mid}}}$ and $\mW_{\text{up}} \in \sR^{D_{\text{mid}} \times  D_{\text{hidden}}}$ are added for the query and value projections along with the original matrix  $\mW_Q$ and $\mW_V \in \sR^{D_{\text{hidden}} \times D_{\text{hidden}}}$:
\begin{equation}
    \mQ = (\mW_Q^\intercal + \alpha \mW_{\text{up}}^\intercal \mW_{\text{down}}^\intercal) \vh_{\text{in}},
\end{equation}
where $\alpha$ is a fixed scalar hyperparameter for scaling the task-specific differences. The form of the trainable matrices in LoRA is quite similar to those in adapter or prefix-tuning, but there is no activation function $\phi$ in between.

\section{Unifying \task Methods}
\subsection{Task Formulation}
Given a large PLM $\gM$ with size $|\gM|$ that cannot be fine-tuned directly due to computational or storage cost, suppose that we have a list of \task methods $\{m_i\}$, the trainable parameters of which are negligible (\ie, $\sum_i{|m_i|} \ll |\gM|$), our goal is to design a unified \task framework that incorporates $\{m_i\}$ as submodules and learns to dynamically activate (upweight) different submodules when appropriate under different scenarios, such that one could achieve satisfactory results in terms of both model effectiveness and robustness without the hassle of permuting all the method$\times$task$\times$data combinations.

\subsection{Proposed Method}
\start{Motivation \& Intuition}
During the analysis of individual \task methods, we observe that different \task methods exhibit diverse characteristics and perform rather differently on the same task.
For example, prefix-tuning generally performs well on natural language inference tasks regardless of the size of training data.
Also, as can be seen in Fig.~\ref{fig_model} and Sec.~\ref{sec_prelim}, different \task methods often involve different parts of the PLM architecture (\eg, before multi-head attention for prefix-tuning and after feedforward layer for adapter), making it feasible to combine multiple \task methods without (directly) interfering with each other.

In light of the two observations above, we propose a unified \task framework, \ours, which takes a hybrid approach by incorporating multiple \task methods as submodules.
At a high level, \ours improves over single \task methods due to two factors.
First, \ours learns to activate (upweight) the submodules that best suit the current task or specific data sample and deactivate (downweight) the rest.
Second, we find that \ours generally performs better than taking the best performance of all its submodules used individually on each task, suggesting that there could be some compounding effects that lead to better model effectiveness when multiple \task methods (that modify different parts of the PLM) are used.

Next, we will introduce how different \task methods can be incorporated into \ours via gating mechanism.

\start{Gating Mechanism}
To achieve fine-grained control of submodule (de)activation, we add a trainable gate $\gG_{m_i}$ for each submodule $m_i \in \{\text{A, P, L}\}$ in every Transformer layer (see Fig.~\ref{fig_model}).
The letters A, P, L stand for Adapter, Prefix-tuning, and LoRA, respectively.
Intuitively, if $m_i$ is useful for a given data $\times$ task setup (or a particular instance), the gate output for $m_i$ would be higher such that $m_i$ plays a more important role.
The actual interplay of submodules, however, is more complicated given the interdependency of the submodules and the compounding effects of multiple layers. 

Specifically, for adapter, there is a residual connection between the feedforward network and the adapter submodule that sums the adapter input (before normalization) $\vh_{F}$ and output $\vh_A$ as its final output: $\vh_A' = \vh_A + \vh_{F}$.
We design a gating function $\gG_A \in (0, 1)$ that estimates the importance of adapter by its direct input $\vh_{FN}$ using a feedforward network with sigmoid activation and then scales its output: $\vh_A' = \gG_A \vh_A + \vh_{F}$.
The adapter submodule is effectively bypassed if $\gG_A \approx 0$.

Similarly, for prefix-tuning, we design a gating function $\gG_P \in (0, 1)$ that is applied to the prefix vectors ($\mP_K$ and $\mP_V$) with the representation of the original tokens ($\mK$ and $\mV$) intact.
In this way, the impact of the prefix would be diminished if the gate output of the prefix-tuning submodule is low.\footnote{Prefix-tuning cannot be fully eliminated as adapter or LoRA due to the softmax operation in multi-head attention.}
The gating function $\gG_P$ is estimated by the Transformer layer input $\vh_{\text{in}}$ with another feedforward network. 

As for LoRA, we note that there is already a \textit{constant} scaling factor $\alpha$ in its original design that resembles the purpose of our gating mechanism.
We thus simply make the factor learnable per layer by a third feedforward network that takes $\vh_{\text{in}}$ as input instead of specifying a constant manually: $\theta_{\text{task}} = \theta_{\text{pre-trained}} + \gG_L \delta_{\text{task}}$.

Despite the seeming simplicity of \ours, we note that it is nontrivial for a unified approach to work well under different scenarios. Naively combining different \task methods as a hybrid approach could lead to mixed or worse performance than using individual methods, as observed in both our experiments and prior studies \cite{hu2021lora}.

\section{Experiments}
We conduct extensive experiments with 8 tasks $\times$ 4 data sizes $\times$ 7 methods $\times$ 5 runs per setup, along with additional analysis for particular methods, resulting in 1,000+ runs in total.

\subsection{Experiment Setup}
\start{Task Setup}
We conduct experiments on the General Language Understanding Evaluation (GLUE) benchmark \cite{wang2019glue}, which involves four types of natural language understanding tasks including linguistic acceptability (CoLA), sentiment analysis (SST-2), similarity and paraphrase tasks (MRPC, STS-B, QQP), and natural language inference (MNLI, QNLI, RTE). We exclude the WNLI dataset following prior studies \cite{houlsby2019parameter,devlin-etal-2019-bert}.

\begin{table*}[t]
\centering
\resizebox{\textwidth}{!}{
\begin{tabular}{l ccccccccc}
\toprule

\textbf{Method} & \textbf{SST-2} & \textbf{MRPC} & \textbf{CoLA} & \textbf{RTE} & \textbf{QNLI} & \textbf{STS-B} & \textbf{MNLI} & \textbf{QQP} &\bf{Avg.}\\

\midrule
\multicolumn{10}{l}{[$K=100$]  Test Performance}\\
Fine-tuning & 79.61$_{4.25}$	 & \underline{81.81}$_{0.35}$	 & \underline{16.56}$_{4.34}$	 & \underline{55.88}$_{1.64}$	 & 69.25$_{5.94}$	 & 74.07$_{6.51}$	 & \underline{42.56}$_{3.43}$	 & 60.41$_{6.42}$	 & 60.02$_{1.84}$	\\
BitFit & 62.94$_{4.85}$	 & 81.09$_{0.17}$	 & 2.71$_{1.57}$	 & 47.65$_{3.56}$	 & 42.46$_{1.37}$	 & 54.53$_{0.56}$	 & 38.16$_{0.53}$	 & 59.56$_{0.39}$	 & 48.64$_{0.78}$	\\
Adapter & 80.48$_{2.94}$	 & 81.40$_{0.19}$	 & 2.02$_{4.04}$	 & 52.78$_{0.27}$	 & 72.25$_{0.49}$	 & 77.32$_{1.54}$	 & 38.81$_{3.64}$	 & 60.88$_{4.00}$	 & 58.24$_{0.99}$	\\
Prefix-tuning & 60.87$_{12.47}$	 & 81.22$_{0.00}$	 & 0.00$_{0.00}$	 & \textbf{55.96}$_{2.00}$	 & 71.91$_{2.69}$	 & 57.69$_{0.02}$	 & 40.58$_{2.49}$	 & 15.68$_{0.12}$	 & 47.99$_{1.77}$	\\
\ \quad \textrightarrow $L=50$ &79.52$_{1.21}$	 & {81.22}$_{0.00}$	 & {5.19}$_{8.62}$	 & {49.24}$_{2.08}$	 & {66.33}$_{2.45}$	 & {7.15}$_{10.37}$	 & {33.66}$_{2.21}$	 & {58.32}$_{3.18}$	 & {47.56}$_{1.37}$	\\
LoRA & \underline{81.56}$_{0.94}$	 & 81.66$_{0.81}$	 & 13.31$_{10.00}$	 & 55.02$_{1.75}$	 & \textbf{73.52}$_{1.20}$	 & 49.35$_{21.87}$	 & 39.60$_{4.98}$	 & 0.09$_{0.02}$	 & 49.26$_{2.19}$	\\
{\cellcolor{na}\oursAP} & 77.22$_{3.75}$	 & \textbf{81.86}$_{0.70}$	 & 14.42$_{10.24}$	 & 55.52$_{2.16}$	 & 72.26$_{0.89}$	 & \underline{79.14}$_{1.97}$	 & \textbf{42.59}$_{1.20}$	 & \textbf{63.41}$_{1.44}$	 & \underline{60.80}$_{1.53}$	\\
{\cellcolor{na}\oursAPL} & \textbf{82.36}$_{0.86}$	 & 81.71$_{0.72}$	 & \textbf{23.62}$_{8.83}$	 & 55.45$_{1.28}$	 & \underline{73.19}$_{0.93}$	 & \textbf{79.37}$_{1.07}$	 & 42.30$_{1.88}$	 & \underline{62.70}$_{2.55}$	 & \textbf{62.59}$_{1.44}$	\\

\midrule

\multicolumn{10}{l}{[$K=500$]  Test Performance}\\
Fine-tuning & \textbf{85.67}$_{0.97}$    & \underline{83.34}$_{0.55}$    & 36.47$_{2.69}$        & 59.64$_{1.10}$        & 77.30$_{0.49}$        & \underline{84.96}$_{1.19}$    & 55.84$_{0.85}$        & \underline{68.23}$_{1.39}$     & 68.93$_{0.65}$       \\
BitFit & 83.44$_{0.63}$  & 82.16$_{0.37}$        & 3.32$_{2.59}$         & 61.88$_{2.75}$        & 69.15$_{9.91}$        & 76.30$_{0.36}$        & 40.82$_{3.30}$        & 65.29$_{3.66}$        & 60.30$_{1.91}$        \\
Adapter & 84.54$_{1.37}$         & 82.53$_{0.36}$        & 38.65$_{3.97}$        & 59.35$_{3.09}$        & 77.39$_{0.84}$        & 83.52$_{0.33}$        & 50.04$_{1.72}$        & 68.12$_{0.95}$        & 68.02$_{0.77}$        \\
Prefix-tuning & 83.65$_{0.69}$   & 82.96$_{1.63}$        & 38.16$_{2.25}$        & 63.18$_{2.70}$        & 78.50$_{1.12}$        & 79.75$_{1.49}$        & \textbf{58.06}$_{1.04}$       & 54.34$_{25.91}$        & 67.32$_{3.42}$       \\
LoRA & \underline{84.98}$_{1.10}$        & 82.53$_{0.70}$        & \textbf{39.86}$_{2.71}$       & 63.03$_{2.57}$        & \textbf{79.46}$_{0.66}$       & 65.05$_{26.31}$       & \underline{56.54}$_{2.05}$     & 55.46$_{27.74}$       & 65.86$_{4.18}$       \\
{\cellcolor{na}\oursAP} & 84.84$_{0.28}$  & 83.25$_{0.51}$        & \underline{39.84}$_{5.01}$    & \underline{63.32}$_{1.72}$    & 78.36$_{1.06}$        & 84.53$_{0.48}$        & 56.08$_{3.26}$        & 68.14$_{1.39}$         & \underline{69.79}$_{1.02}$   \\
{\cellcolor{na}\oursAPL} & 84.91$_{1.41}$         & \textbf{83.56}$_{0.59}$       & 39.81$_{2.55}$        & \textbf{64.12}$_{2.45}$       & \underline{79.28}$_{0.63}$    & \textbf{85.26}$_{0.70}$       & 54.07$_{3.74}$         & \textbf{68.87}$_{0.41}$       & \textbf{69.98}$_{0.42}$      \\

\midrule

\multicolumn{10}{l}{[$K=1000$]  Test Performance}\\
Fine-tuning & \underline{86.54}$_{1.01}$         & 84.87$_{0.64}$        & 43.26$_{2.60}$        & 62.31$_{2.10}$        & 79.03$_{1.11}$        & 86.39$_{0.34}$        & 61.95$_{1.20}$        & \textbf{71.09}$_{0.77}$        & 71.93$_{0.37}$       \\
BitFit & 83.99$_{0.39}$  & 83.95$_{0.81}$        & 22.44$_{17.10}$       & 62.89$_{1.40}$        & 77.43$_{0.53}$        & 79.04$_{0.61}$        & 52.87$_{0.72}$        & 69.50$_{0.16}$        & 66.51$_{2.22}$        \\
Adapter & 85.60$_{0.63}$         & 84.49$_{0.60}$        & 42.33$_{1.98}$        & 61.81$_{1.57}$        & 79.68$_{0.23}$        & 85.52$_{0.29}$        & 57.86$_{2.44}$        & 70.32$_{0.71}$        & 70.95$_{0.55}$        \\
Prefix-tuning & 85.09$_{0.99}$   & 83.66$_{1.82}$        & 44.07$_{2.90}$        & \textbf{66.71}$_{2.72}$       & 80.34$_{0.70}$        & 82.38$_{1.25}$        & \textbf{63.59}$_{1.12}$       & 68.58$_{0.35}$         & 71.81$_{0.52}$       \\
LoRA & 86.26$_{1.22}$    & \underline{86.04}$_{0.99}$    & \textbf{45.50}$_{1.11}$       & \underline{65.63}$_{2.11}$    & \underline{81.00}$_{0.98}$    & 81.56$_{1.97}$        & 61.32$_{1.65}$        & 70.89$_{0.81}$         & 72.28$_{0.69}$       \\
\cellcolor{na}\oursAP & 86.17$_{0.37}$  & 85.86$_{1.05}$        & 44.33$_{3.55}$        & 64.91$_{1.92}$        & 80.65$_{0.57}$        & \underline{86.82}$_{0.23}$    & 62.17$_{0.99}$        & 69.95$_{0.90}$ & \underline{72.61}$_{0.53}$   \\
\cellcolor{na}\oursAPL & \textbf{87.06}$_{0.81}$        & \textbf{86.65}$_{1.10}$       & \underline{45.44}$_{1.97}$    & 65.49$_{1.92}$        & \textbf{81.22}$_{0.51}$       & \textbf{87.10}$_{0.21}$       & \underline{62.49}$_{0.94}$     & \underline{70.99}$_{0.95}$    & \textbf{73.31}$_{0.52}$      \\

\bottomrule
\end{tabular}}
\upv
\caption{Results on the GLUE benchmark with $K=\{100, 500, 1000\}$ training samples. The evaluation metrics are Matthew's Correlation for CoLA, F1 for MRPC and QQP, Spearman's correlation for STS-B, and accuracy for the rest. For MNLI, we evaluate on the matched dataset. We report average performance on five random seeds with standard deviation as the subscript. \textbf{Best} and \underline{2nd best} methods under each setup are bold and underlined.} 
\label{table_glue_low_major}
\downv
\end{table*}

\start{Data Setup}
We mainly consider a low-resource setting where training data is limited and the performance of different methods varies much. We sample a small subset of the training set for each task with size $K=\{100, 500, 1000 \}$.
As it is infeasible to submit considerable runs to the GLUE leaderboard (2 submissions/day), we take 1,000 samples on the training set as the development set to select the best checkpoint and use the original development set as the test set.
To reduce variance, we shuffle the data with 5 random seeds and report the average performance.
Additionally, we consider a high-resource setting where the whole training set is used and the best performance on the GLUE development set is reported.

\start{Compared Methods}
We mainly compare \ours with fine-tuning and four representative \task methods: adapter \cite{houlsby2019parameter}, prefix-tuning \cite{li-liang-2021-prefix}, BitFit \cite{ben2021bitfit}, and LoRA \cite{hu2021lora}.
For completeness, we consider two model variants \oursAP and \oursAPL, which incorporate 2 and 3 \task methods, respectively.

\start{Implementation Details}
We use BERT$_\text{base}$ \cite{devlin-etal-2019-bert} as the base model in the experiments. 
Consistent results are observed in our preliminary experiments with BART$_\text{large}$ \cite{lewis-etal-2020-bart} (provided in App.~\ref{sec_bart_res}).
We implement and evaluate all the methods in the same codebase to ensure a fair comparison.
We largely follow the default hyperparameters of different methods and keep them the same on all the tasks for generalizability.
We set the prefix length $L=10$, adapter bottleneck size $D_{\text{mid}}=48$, LoRA rank $D_{\text{mid}}=8$ if not specified otherwise.\footnote{While these hyperparameters may lead to differences in trainable parameters, we keep them for analysis as they are used by the official implementation. Also, we observe that more trainable parameters do not guarantee better results.}
More implementation and hyperparameter details can be found in App.~\ref{sec_implementation}.

\subsection{Analysis of Individual \task Methods}
\label{sec_exp_existing}
In Table~\ref{table_glue_low_major}, we show the performance of different methods on the GLUE benchmark with various sizes of training data.
The results on the development sets are generally consistent with the test sets and provided in App.~\ref{sec_additional_res}.
Although the average performance of different methods over 8 tasks is sometimes similar, the differences between tasks are quite significant under certain setups and can be as large as 5\textasciitilde9 points on a specific task (\eg, STS-B and MNLI, $K=500$) even when excluding cases where some methods fail to learn effectively (\eg, prefix-tuning on QQP, $K=100$).

Next, we will analyze and examine each individual \task method more closely.

\start{Analysis of Adapter}
The performance of adapter is relatively stable -- there is no significantly better or worse result than fine-tuning consistent on different tasks or sizes of training data.
In general, adapter is slightly worse than fine-tuning in most cases.
We do not observe that adapter consistently outperforms fine-tuning in the low-resource setting as in \citet{he-etal-2021-effectiveness}, possibly because they tune model hyperparameters on each task, which could be computationally prohibitive when there are considerable tasks.
For example, they choose the bottleneck size $D_{\text{mid}}$ from \{64, 128, 256\}, while $D_{\text{mid}}=48$ is fixed across tasks in our experiments.
Also, we only add one adapter in each Transformer layer instead of two following \citet{pfeiffer-etal-2021-adapterfusion}.
These two differences result in 62.4\%\textasciitilde90.5\% fewer parameters than the adapter used in \citet{he-etal-2021-effectiveness}.

\begin{figure}[ht]
    \centering
    \includegraphics[width=0.81\linewidth]{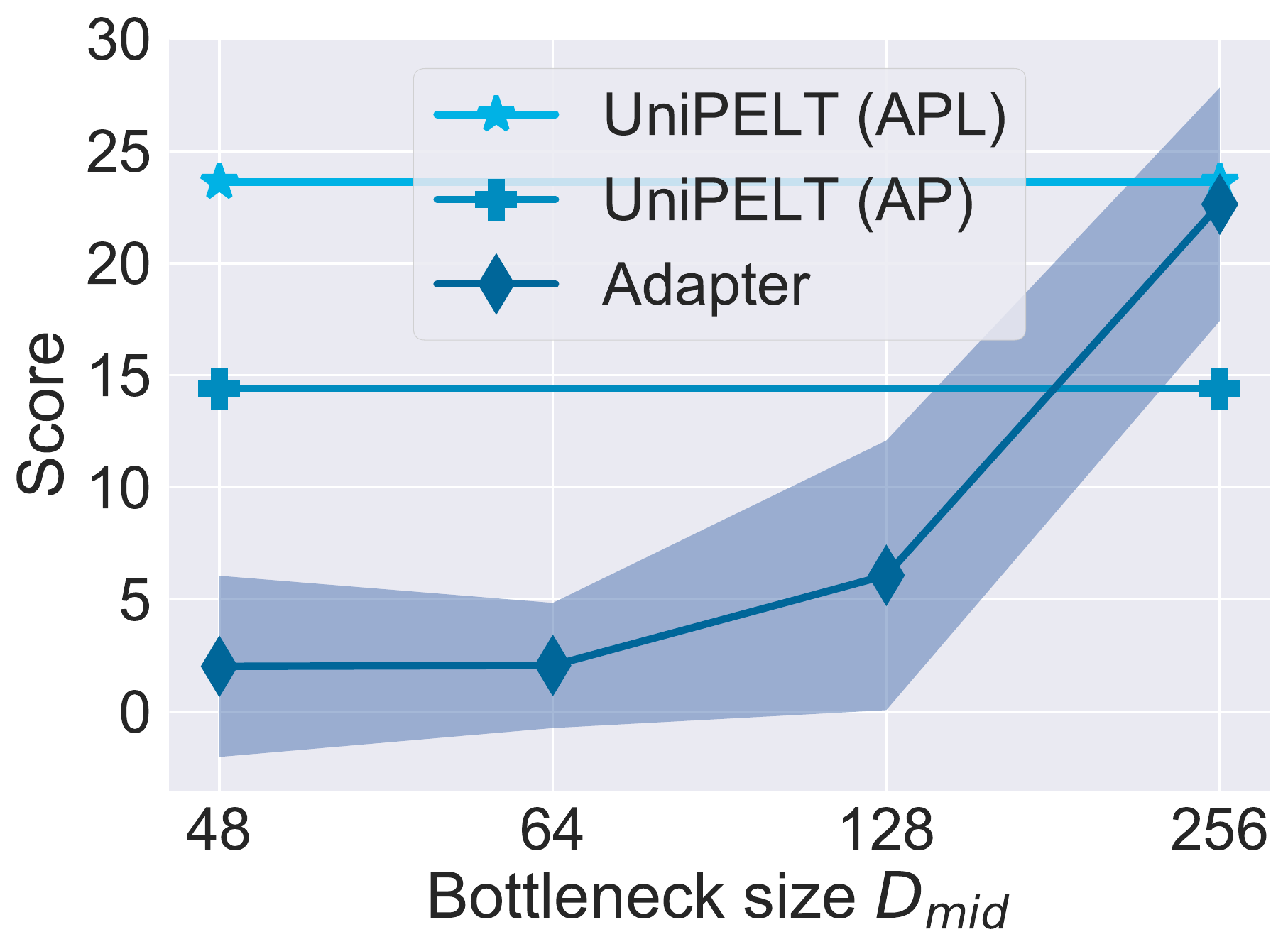}
    \upv
    \caption{Performance changes when the bottleneck size of adapter is increased (on CoLA, $K=100$).}
    \label{fig_adapter_bottleneck}
    \downv
\end{figure}
To further study the effect of bottleneck size $D_{\text{mid}}$ in adapter, we increase $D_{\text{mid}}$ and re-evaluate adapter on a setup that it performs poorly (CoLA, $K=100$).
As shown in Fig.~\ref{fig_adapter_bottleneck}, the performance of adapter is increased gradually and becomes significantly better only when $D_{\text{mid}}=256$, which involves 5.3$\times$ trainable parameters than the adapter used originally ($D_{\text{mid}}=48$), 4.3$\times$ than \oursAP, and 3.4$\times$ than \oursAPL, suggesting that a larger bottleneck size could be beneficial when adapter learns ineffectively. 

On the other hand, there are certain tasks (\eg, STS-B) that adapter largely outperforms competitive methods such as prefix-tuning and LoRA regardless of the size of training data, suggesting that one should favor adapter over other \task methods under certain scenarios as well.

\start{Analysis of Prefix-tuning}
Prefix-tuning performs poorly with $K=\{100, 500 \}$ and becomes on par with fine-tuning when $K$ reaches 1000.
We also observe that prefix-tuning fails to learn effectively on certain tasks when the training data is limited (\eg, $K=100$ on SST-2 and $K=500$ on QQP), leading to unsatisfactory performance and (or) large variance across different runs.
Similar phenomena have been observed in a concurrent study \cite{gu2021ppt} on few-shot prompt-tuning.

To ensure that the poor performance of prefix-tuning is not due to its fewer trainable parameters (based on its default setting), we further increase the prefix length to $L=50$ such that its trainable parameters are comparable to adapter, and re-evaluate prefix-tuning on all 8 tasks with $K = 100$.
For the 4 tasks where prefix-tuning ($L=10$) performs poorly (SST2, CoLA, STS-B, and QQP), while its performance is significantly improved on 3 tasks, it also performs significantly worse on the other task (STS-B), which suggests that training instability in the low-resource regime is still an issue for prefix-tuning even with more trainable parameters.\footnote{Tuning other hyperparameters like learning rate does not appear to alleviate the issue either.}
Besides, prefix-tuning ($L=50$) still lags behind adapter or \oursAP on 3 of the 4 tasks.
Furthermore, the average performance of prefix-tuning ($L=50$) on 8 tasks is even slightly worse than with $L=10$, which indicates that increasing prefix length may not be a panacea for all the scenarios. A larger $L$ also leads to significant training/inference slowdown due to the costly multi-head attention.
More broadly, such results suggest that {using more trainable parameters does not guarantee better performance}.

On the bright side, prefix-tuning performs well on certain tasks such as natural language inference (RTE and MNLI) with various sizes of training data, which suggests that one should also prefer prefix-tuning in certain cases.

\start{Analysis of BitFit \& LoRA}
Tuning only the bias terms of the model does not lead to very satisfactory results in our experiments -- BitFit never performs the best and generally performs the worst in different data and task setups.
Therefore, we do not consider BitFit in the following experiments and exclude BitFit as a submodule of \ours.
As for LoRA, there are a few setups where LoRA fails to learn effectively as well, such as STS-B and QQP ($K=\{100, 500\}$), leading to high variance across runs.
Apart from that, LoRA performs quite competitively despite using fewer trainable parameters than methods like adapter, especially when $K=1000$, achieving the best or 2nd best performance on 4 of 8 tasks.

\begin{figure}[t]
    \centering
    \includegraphics[width=0.93\linewidth]{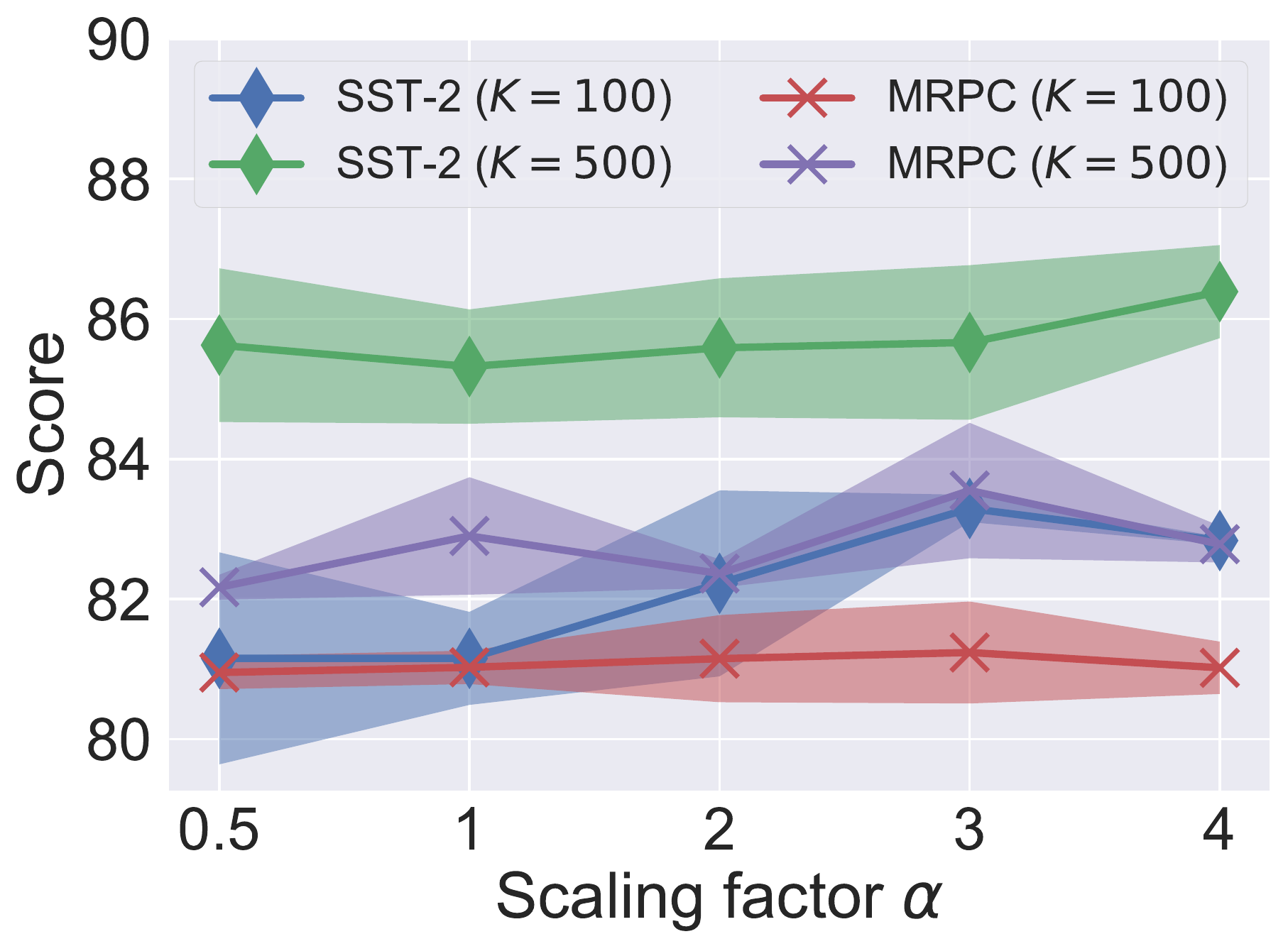}
    \upv
    \caption{Performance comparison of various scaling factors for LoRA on 2$\times$2 task and data setups.}
    \label{lora_alpha}
    \downv
\end{figure}

As LoRA has a scaling factor $\alpha$ that can be seen as a static gating function under our formulation, we further investigate its importance by evaluating LoRA with different $\alpha$.
As shown in Fig.~\ref{lora_alpha}, LoRA is quite sensitive to the scaling factor and there seems to be no single optimal value that works well across multiple task and data setups.
Such findings suggest that gating is critical and motivate us to use more fine-grained and dynamic control for \ours.
Besides, we observe that increasing $\alpha$ consistently results in faster convergence, possibly because the trainable parameters would receive larger gradient updates with a larger $\alpha$.

\subsection{Analysis of \ours}
Next, we will turn to our proposed framework \ours, which incorporates multiple existing \task methods as submodules.

\start{Low-Resource Performance}
Overall, \oursAPL and \oursAP consistently achieve the best and second best average performance on both the development and test sets regardless of the number of training samples.
The gains are generally 1\textasciitilde4\% over the submodule that performs the best (when used individually).
Such results demonstrate the advantages of our hybrid approach regarding model effectiveness and generalizability.

At the per-task level, \oursAPL and \oursAP perform the best or second best on 7/6/7 of 8 tasks when trained with 100/500/1,000 samples, and never perform the worst in any setup.
When comparing the two variants, \oursAPL outperforms \oursAP on 4/6/8 of 8 tasks when trained with 100/500/1,000 samples.
Such results indicate that \ours is quite robust and performs reliably under different scenarios.
The improvements of \ours over its submodules are generally larger when having fewer training samples, suggesting that \ours performs especially well in the low-resource regime.
In particular, on the tasks where other PELT methods fail to learn effectively such as CoLA and QQP ($K=100$), \ours manages to achieve performance better than fine-tuning.

\start{\ours vs. Upper Bound}
In Table~\ref{table_take_best}, we show the comparison of \ours and the upper bound that takes the best performance of its submodules on each task.
We observe that both \oursAP and \oursAPL perform similarly or even better than their upper bound, which suggests that \ours successfully learns to leverage different submodules and maintains (near) optimal performance under different setups.
The fact that \ours can outperform the upper bound also hints that a mixture of \task methods (involving different parts of the PLM) might be inherently more effective than single methods (with a limited scope of the PLM architecture).

\begin{table}[ht]
\centering
\resizebox{1\columnwidth}{!}{
\begin{tabular}{lrr|rr}
\toprule

$K$ & $\max(\{A, P\})$ & \ours & $\max(\{A, P, L\})$ & \ours \\
\midrule
100 & 58.86 & \textbf{60.80} & 60.60 & \textbf{62.59}\\
500 & 69.69 & \textbf{69.79} & \textbf{70.02} & 69.98\\
1000 & 72.58 & \textbf{72.61} & 73.19 & \textbf{73.31}\\

\bottomrule
\end{tabular}}
\upv
\caption{Comparison of average test performance between \ours and the upper bound that takes the best performance of its submodules on each task. } 
\label{table_take_best}
\downv
\end{table}

\begin{table*}[t]
\vspace{-0.2cm}
\centering
\resizebox{.87\textwidth}{!}{
\begin{tabular}{l ccccccccc}
\toprule

\textbf{Method} & \textbf{SST-2} & \textbf{MRPC} & \textbf{CoLA} & \textbf{RTE} & \textbf{QNLI} & \textbf{STS-B} & \textbf{MNLI} & \textbf{QQP} &\bf{Avg.}\\

\midrule
\multicolumn{10}{l}{[$K=all$]  Best Performance on GLUE Dev}\\
Fine-tuning & 91.63      & \underline{90.94}     & \textbf{62.08}        & 66.43         & 89.95         & \textbf{89.76}        & 83.23         & \textbf{87.35}        & 82.67        \\
Adapter & \textbf{91.86}         & 89.86         & 61.51         & 71.84         & \underline{90.55}     & 88.63         & 83.14         & {86.78}     & 83.02        \\
Prefix-tuning & 90.94    & \textbf{91.29}        & 55.37         & \textbf{76.90}        & 90.39         & 87.19         & 81.15         & 83.30         & 82.07        \\
LoRA & 91.51     & 90.03         & 60.47         & 71.48         & 89.93         & 85.65         & 82.51         & 85.98         & 82.20        \\
\cellcolor{na}\oursAP & \textbf{91.86}      & 90.28         & 61.15         & 71.84         & \textbf{90.77}        & 88.86         & \underline{83.41}     & 86.74         & \underline{83.12}    \\
\quad -NoGate & \underline{91.74}   & 90.18         & 58.63         & 71.12         & 90.30         & 88.76         & 81.58         & 85.53         & 82.23        \\
\cellcolor{na}\oursAPL & 91.51  & \underline{90.94}        & \underline{61.53}     & \underline{73.65}     & 90.50         & \underline{88.93}     & \textbf{83.89}        & \underline{87.12}     & \textbf{83.50}       \\

\bottomrule
\end{tabular}}
\upv
\caption{Results on the GLUE benchmark when all training samples are used.} 
\label{table_glue_all}
\downv
\end{table*}

\start{High-Resource Performance}
In Table~\ref{table_glue_all}, we list the performance of different methods when all training samples are used.
\ours again achieves the best overall performance. The gains are not as significant as in the low-resource setting, which is somewhat expected as existing \task methods typically perform on par with fine-tuning given abundant training data and the potential of improvement is not as high.
That said, the performance of \ours is still the best or 2nd best on all 8 tasks, and generally comparable to the best submodule used individually on each task.
Besides, simply combining multiple \task methods without gating does not work well in the high-resource setting -- although \ours-NoGate never performs the worst in each task, its average performance is unsatisfactory (-0.89 vs. \ours).

\subsection{Efficiency of \task Methods}
We benchmark the efficiency of \task methods and list in Table~\ref{tab_param} their number of trainable parameters and training/inference time relative to fine-tuning.

\start{Parameter Efficiency}
As the trainable parameters in \task methods are almost negligible, combining multiple methods does not lead to significant losses in parameter efficiency. \ours still has few trainable parameters compared to fine-tuning (0.99\%\textasciitilde1.26\%).
The parameters can be further reduced if one uses more parameter-efficient variants (\eg, \citet{mahabadi2021compacter}), which can be easily swapped with the vanilla version used in our current framework.
Also, note that more trainable parameters do not always lead to better performance, as shown in our experiments and prior studies \cite{he-etal-2021-effectiveness,pfeiffer-etal-2021-adapterfusion}.

\begin{table}[t]
\centering
\resizebox{1\columnwidth}{!}{
\begin{tabular}{lrrr}
\toprule

\textbf{Method} & \textbf{\#Param.} & \textbf{Time$_T$} & \textbf{Time$_I$}\\
\midrule

Fine-tuning & 110M (100\%) & 100\% & 100\%\\
BitFit & 103K (0.09\%) & 65\% & 102\% \\
Prefix-tuning  & 184K (0.17\%) & 56\% & 114\% \\
LoRA & 295K (0.27\%) & 53\% & 105\% \\
Adapter  & 895K (0.81\%) & 55\% & 107\%  \\
\cellcolor{na}\oursAP &  1.1M (0.99\%) & 55\% & 118\%\\
\cellcolor{na}\oursAPL & 1.4M (1.26\%) & 67\% & 127\% \\

\bottomrule
\end{tabular}}
\upv
\caption{Number of trainable parameters and $T$raining/$I$nference time relative to fine-tuning.  } 
\label{tab_param}
\downv
\end{table}

\start{Training and Inference Efficiency}
Due to parameter efficiency, all \task methods train 30\%\textasciitilde50\% faster than fine-tuning and incorporating multiple \task methods into \ours does not suffer from slower training.
On the other hand, the inference time of PELT methods is generally longer since they involve more FLOPs. \ours has a slightly larger inference overhead (4\%\textasciitilde11\% compared to its slowest submodule), which we argue is insignificant since larger models that may achieve similar performance gains (\eg, BERT$_\text{large}$) need around 300\% inference time \cite{hf_benchmark}.

\section{Related Work}
\start{Parameter-Efficient Tuning of PLMs}
As it is increasingly infeasible to train and store full copies of large PLMs for various downstream tasks, how to efficiently tune the PLMs with few trainable parameters becomes critical.
Existing \task methods can be largely divided into two categories based on whether new trainable parameters are introduced.
Specifically, one may either train a subset of the model parameters such as the prediction head \cite{lee2019would} and bias terms \cite{ben2021bitfit}, or introduce task-specific parameters to different parts of the PLM such as before multi-head attention \cite{li-liang-2021-prefix} or after feedforward layer \cite{houlsby2019parameter}.
As the number of \task methods keeps increasing, the purpose of \ours is to better understand and leverage the distinctions of various methods instead of proposing yet another method.

\start{Mixture-of-Experts}
\ours is also related to approaches that involve a high-capacity network and activate (upweight) different parts of the network given different inputs.
One notable example is Mixture-of-Experts (MoE) \cite{jacobs1991adaptive,shazeer2017outrageously}, which maintains a set of experts (neural networks) and one or more trainable gates that select a combination of the experts specific to each input.
Despite being conceptually similar, \ours is different from MoE: the submodules in \ours are not combined explicitly by summation like MoE but in sequential order and affect each other implicitly. Moreover, the ``experts'' are diverse in \ours while usually homogeneous or identical in MoE methods.

\section{Conclusion}
In this paper, we present a comprehensive study of representative parameter-efficient language model tuning (\task) methods and propose a unified framework, which incorporates different \task methods as submodules and learns to activate the most appropriate submodules for a given task or data setup.
Our proposed framework consistently outperforms conventional fine-tuning as well as the submodules that it incorporates under different setups, and generally surpasses the upper bound that takes the best performance of each submodule used individually on each task.
Our findings suggest that a mixture of multiple \task methods that involve different parts of the PLM may be favorable regarding both model effectiveness and robustness.
For future work, we will try to better understand the discrepancy of various \task methods in different scenarios. We also plan to investigate a multi-task setting where multiple submodules can be activated and cooperate at the task level.

\section*{Acknowledgements}
We thank Xiang Lisa Li, Hai Ye, Rabeeh Karimi Mahabadi, Junxian He, Yiqing Xie, Yaqing Wang, and Liyuan Liu for helpful discussions and feedback.
We thank anonymous reviewers for valuable comments and suggestions.

\bibliography{anthology,0-custom}
\bibliographystyle{acl_natbib}

\newpage
\clearpage

\appendix

\section{Prefix-tuning vs. Prompt-based Fine-tuning}
\label{sec_prefix_vs_prompt}
We note that prefix-tuning (or prompt-tuning) is different from prompt-based \textit{fine-tuning} methods~\cite{schick-schutze-2021-exploiting,gao-etal-2021-making} in many ways:
(1) Prompt-based fine-tuning is not parameter-efficient as it updates all model parameters while prefix-tuning only updates the prefix matrix $\mP$.
(2) The prompts are only used in model input for prompt-based fine-tuning but added to \textit{every} Transformer layer in prefix-tuning (stored as different vectors).
(3) Prompt-based fine-tuning typically leverages carefully designed \textit{natural language} prompts while prefix-tuning uses continuous  prompts (virtual tokens).

\section{Implementation Details}
\label{sec_implementation}

\start{Data Preparation}
We shuffle the training set with seed $s$, take the first $K$ samples as the new training set, and the next 1,000 samples as the development set.
We use $s=\{111, 222, 333, 444, 555 \}$ as the data seeds and the same seed ($s=42$) for model training. We also conduct another set of preliminary experiments by fixing the data and using 5 different random seeds for model training, the results of which are similar (Table~\ref{table_bart}).

\start{Hyperparameters}
We adopt AdapterHub \cite{pfeiffer2020AdapterHub}, a library based on HuggingFace Transformers \cite{wolf2019huggingface},  as our codebase.
We largely follow the recommended hyperparameters used in different methods for a fair comparison.
We set the input length to 128 and the training batch size to 16.
We set the number of epochs to 50 and adopt early stopping with a patience of 10 non-increasing epochs.
We set the learning rate of fine-tuning and adapter to 2e-5 and 1e-4 according to the findings in prior studies \cite{pfeiffer2020AdapterHub,he-etal-2021-effectiveness}. 
For prefix-tuning and \ours, as they are not previously evaluated on NLU tasks, we tune their learning rates from \{1e-4, 2e-4, 5e-4\} on the development set and set to 2e-4 and 5e-4, respectively.
For BitFit and LoRA, we choose the learning rates commonly used in their own experiments (1e-4 and 5e-4, respectively).
We set $\alpha=2$ and $r=8$ in LoRA according to its official scripts.

\section{BART Results}
\label{sec_bart_res}

In our preliminary experiments, we also evaluated \ours on BART$_\text{large}$ \cite{lewis-etal-2020-bart}.
We show the results of fine-tuning, adapter, prefix-tuning, and \oursAP in Table~\ref{table_bart}.
1000 training examples are used and the average best performance on the GLUE development set is reported (excluding QQP). 
The results are largely consistent with those on BERT$_\text{base}$. \ours again achieves the best performance with notably smaller variance.

\begin{table}[t]
\centering
\resizebox{1\columnwidth}{!}{
\begin{tabular}{l cccc}
\toprule

\textbf{Setup} & \textbf{Fine-tuning} & \textbf{Adapter} & \textbf{Prefix-tuning} & \textbf{\oursAP} \\

\midrule
Model seed & 78.39$_{2.92}$ & 77.12$_{0.50}$ & 73.16$_{2.89}$ &  \textbf{78.66}$_{0.24}$ \\
Data seed & 77.55$_{2.94}$ & 76.87$_{0.55}$ & 71.90$_{2.47}$ & \textbf{79.02}$_{0.44}$\\

\bottomrule
\end{tabular}}
\upv
\caption{Average performance with $K=1000$ on the GLUE benchmark with BART$_\text{large}$ as the base model. Results are averaged over 5 runs by changing the model or data seeds.} 
\label{table_bart}
\downv
\end{table}

\section{Detailed Performance}
\label{sec_additional_res}
In Table~\ref{table_glue_lowres_detail}, we list the detailed results on both development and test sets of the GLUE benchmark. The observations and findings are largely consistent on the two evaluation splits.

\begin{table*}[t]
\centering
\resizebox{\textwidth}{!}{
\begin{tabular}{l ccccccccc}
\toprule

\textbf{Method} & \textbf{SST-2} & \textbf{MRPC} & \textbf{CoLA} & \textbf{RTE} & \textbf{QNLI} & \textbf{STS-B} & \textbf{MNLI} & \textbf{QQP} &\bf{Avg.}\\

\midrule
\multicolumn{10}{l}{[$K=100$] Dev Performance}\\
Fine-tuning & 81.24$_{0.98}$	 & \textbf{81.46}$_{0.78}$	 & \underline{16.94}$_{2.38}$	 & \textbf{58.08}$_{1.63}$	 & 69.66$_{5.03}$	 & 60.64$_{6.97}$	 & \underline{43.18}$_{3.13}$	 & 61.63$_{6.30}$	 & 59.10$_{1.87}$	\\
BitFit & 62.06$_{4.62}$	 & 80.66$_{0.39}$	 & 5.73$_{1.46}$	 & 50.26$_{0.91}$	 & 42.02$_{2.29}$	 & 31.18$_{2.47}$	 & 38.40$_{0.84}$	 & 61.55$_{0.52}$	 & 46.48$_{0.66}$	\\
Adapter & 80.60$_{0.85}$	 & \underline{81.11}$_{0.78}$	 & 2.19$_{4.38}$	 & 53.16$_{1.99}$	 & 72.58$_{0.66}$	 & 66.00$_{3.66}$	 & 40.30$_{2.82}$	 & 62.32$_{3.20}$	 & 57.28$_{0.60}$	\\
Prefix-tuning & 66.24$_{12.03}$	 & 80.51$_{0.31}$	 & 0.00$_{0.00}$	 & 56.60$_{1.25}$	 & 71.94$_{2.58}$	 & 42.81$_{1.93}$	 & 42.26$_{1.89}$	 & 15.14$_{0.95}$	 & 46.94$_{1.43}$	\\
LoRA & \underline{82.54}$_{0.84}$	 & 80.82$_{0.50}$	 & 14.38$_{8.57}$	 & 56.62$_{2.01}$	 & \textbf{74.26}$_{0.89}$	 & 47.87$_{14.05}$	 & 41.38$_{4.59}$	 & 0.00$_{0.00}$	 & 49.73$_{1.29}$	\\
{\cellcolor{na}\oursAP} & 80.40$_{1.95}$	 & 81.02$_{0.54}$	 & 15.07$_{6.46}$	 & 57.68$_{1.63}$	 & 73.50$_{0.54}$	 & \underline{68.19}$_{3.97}$	 & \textbf{44.50}$_{1.11}$	 & \textbf{64.89}$_{0.86}$	 & \underline{60.66}$_{1.16}$	\\
{\cellcolor{na}\oursAPL} & \textbf{83.08}$_{0.54}$	 & 81.08$_{0.53}$	 & \textbf{23.52}$_{5.71}$	 & \underline{57.96}$_{1.49}$	 & \underline{74.00}$_{0.46}$	 & \textbf{68.29}$_{3.01}$	 & 43.10$_{1.13}$	 & \underline{63.41}$_{2.93}$	 & \textbf{61.80}$_{0.77}$	\\

\cmidrule[0.01em]{2-10}
\multicolumn{10}{l}{[$K=100$]  Test Performance}\\
Fine-tuning & 79.61$_{4.25}$	 & \underline{81.81}$_{0.35}$	 & \underline{16.56}$_{4.34}$	 & \underline{55.88}$_{1.64}$	 & 69.25$_{5.94}$	 & 74.07$_{6.51}$	 & \underline{42.56}$_{3.43}$	 & 60.41$_{6.42}$	 & 60.02$_{1.84}$	\\
BitFit & 62.94$_{4.85}$	 & 81.09$_{0.17}$	 & 2.71$_{1.57}$	 & 47.65$_{3.56}$	 & 42.46$_{1.37}$	 & 54.53$_{0.56}$	 & 38.16$_{0.53}$	 & 59.56$_{0.39}$	 & 48.64$_{0.78}$	\\
Adapter & 80.48$_{2.94}$	 & 81.40$_{0.19}$	 & 2.02$_{4.04}$	 & 52.78$_{0.27}$	 & 72.25$_{0.49}$	 & 77.32$_{1.54}$	 & 38.81$_{3.64}$	 & 60.88$_{4.00}$	 & 58.24$_{0.99}$	\\
Prefix-tuning & 60.87$_{12.47}$	 & 81.22$_{0.00}$	 & 0.00$_{0.00}$	 & \textbf{55.96}$_{2.00}$	 & 71.91$_{2.69}$	 & 57.69$_{0.02}$	 & 40.58$_{2.49}$	 & 15.68$_{0.12}$	 & 47.99$_{1.77}$	\\
\ \quad \textrightarrow $L=50$ &79.52$_{1.21}$	 & {81.22}$_{0.00}$	 & {5.19}$_{8.62}$	 & {49.24}$_{2.08}$	 & {66.33}$_{2.45}$	 & {7.15}$_{10.37}$	 & {33.66}$_{2.21}$	 & {58.32}$_{3.18}$	 & {47.56}$_{1.37}$	\\
LoRA & \underline{81.56}$_{0.94}$	 & 81.66$_{0.81}$	 & 13.31$_{10.00}$	 & 55.02$_{1.75}$	 & \textbf{73.52}$_{1.20}$	 & 49.35$_{21.87}$	 & 39.60$_{4.98}$	 & 0.09$_{0.02}$	 & 49.26$_{2.19}$	\\
{\cellcolor{na}\oursAP} & 77.22$_{3.75}$	 & \textbf{81.86}$_{0.70}$	 & 14.42$_{10.24}$	 & 55.52$_{2.16}$	 & 72.26$_{0.89}$	 & \underline{79.14}$_{1.97}$	 & \textbf{42.59}$_{1.20}$	 & \textbf{63.41}$_{1.44}$	 & \underline{60.80}$_{1.53}$	\\
{\cellcolor{na}\oursAPL} & \textbf{82.36}$_{0.86}$	 & 81.71$_{0.72}$	 & \textbf{23.62}$_{8.83}$	 & 55.45$_{1.28}$	 & \underline{73.19}$_{0.93}$	 & \textbf{79.37}$_{1.07}$	 & 42.30$_{1.88}$	 & \underline{62.70}$_{2.55}$	 & \textbf{62.59}$_{1.44}$	\\

\midrule
\multicolumn{10}{l}{[$K=500$]  Dev Performance}\\
Fine-tuning & \underline{86.66}$_{1.40}$         & 82.56$_{0.88}$        & 37.47$_{3.06}$        & 62.88$_{1.79}$        & 77.58$_{1.64}$        & 77.34$_{2.03}$        & 58.50$_{1.53}$        & \underline{69.40}$_{1.32}$     & 69.05$_{0.38}$       \\
BitFit & 84.66$_{1.28}$  & 81.80$_{0.96}$        & 5.66$_{1.87}$         & 61.88$_{0.95}$        & 69.32$_{8.90}$        & 59.55$_{1.41}$        & 42.62$_{3.23}$        & 66.06$_{2.99}$        & 58.94$_{1.65}$        \\
Adapter & 85.74$_{1.03}$         & 82.74$_{0.87}$        & 38.22$_{4.14}$        & 63.52$_{1.98}$        & 78.20$_{1.64}$        & 76.15$_{1.18}$        & 51.30$_{2.65}$        & 69.23$_{1.30}$        & 68.14$_{0.66}$        \\
Prefix-tuning & \textbf{86.72}$_{1.46}$  & 82.26$_{1.16}$        & 40.25$_{5.45}$        & \textbf{66.08}$_{0.83}$       & 78.44$_{1.48}$        & 71.41$_{2.30}$        & \textbf{60.70}$_{1.47}$       & 54.47$_{25.86}$        & 67.54$_{3.45}$       \\
LoRA & 86.36$_{1.37}$    & 82.38$_{1.35}$        & \underline{42.60}$_{3.13}$    & \underline{65.46}$_{1.74}$    & \underline{79.34}$_{1.23}$    & 60.58$_{16.76}$       & 58.70$_{2.17}$        & 56.39$_{28.20}$        & 66.48$_{4.02}$       \\
{\cellcolor{na}\oursAP} & 86.26$_{1.90}$  & \underline{82.77}$_{1.09}$    & 42.48$_{3.38}$        & 65.08$_{1.65}$        & 78.86$_{1.45}$        & \underline{77.83}$_{1.29}$    & \underline{59.46}$_{3.71}$    & 68.95$_{2.14}$         & \underline{70.21}$_{0.78}$   \\
{\cellcolor{na}\oursAPL} & 86.10$_{1.28}$         & \textbf{83.16}$_{0.92}$       & \textbf{43.83}$_{4.73}$       & 64.02$_{2.99}$        & \textbf{79.56}$_{1.49}$       & \textbf{78.54}$_{1.95}$       & 57.08$_{3.87}$         & \textbf{69.56}$_{0.89}$       & \textbf{70.23}$_{0.55}$      \\

\cmidrule[0.01em]{2-10}
\multicolumn{10}{l}{[$K=500$]  Test Performance}\\
Fine-tuning & \textbf{85.67}$_{0.97}$    & \underline{83.34}$_{0.55}$    & 36.47$_{2.69}$        & 59.64$_{1.10}$        & 77.30$_{0.49}$        & \underline{84.96}$_{1.19}$    & 55.84$_{0.85}$        & \underline{68.23}$_{1.39}$     & 68.93$_{0.65}$       \\
BitFit & 83.44$_{0.63}$  & 82.16$_{0.37}$        & 3.32$_{2.59}$         & 61.88$_{2.75}$        & 69.15$_{9.91}$        & 76.30$_{0.36}$        & 40.82$_{3.30}$        & 65.29$_{3.66}$        & 60.30$_{1.91}$        \\
Adapter & 84.54$_{1.37}$         & 82.53$_{0.36}$        & 38.65$_{3.97}$        & 59.35$_{3.09}$        & 77.39$_{0.84}$        & 83.52$_{0.33}$        & 50.04$_{1.72}$        & 68.12$_{0.95}$        & 68.02$_{0.77}$        \\
Prefix-tuning & 83.65$_{0.69}$   & 82.96$_{1.63}$        & 38.16$_{2.25}$        & 63.18$_{2.70}$        & 78.50$_{1.12}$        & 79.75$_{1.49}$        & \textbf{58.06}$_{1.04}$       & 54.34$_{25.91}$        & 67.32$_{3.42}$       \\
LoRA & \underline{84.98}$_{1.10}$        & 82.53$_{0.70}$        & \textbf{39.86}$_{2.71}$       & 63.03$_{2.57}$        & \textbf{79.46}$_{0.66}$       & 65.05$_{26.31}$       & \underline{56.54}$_{2.05}$     & 55.46$_{27.74}$       & 65.86$_{4.18}$       \\
{\cellcolor{na}\oursAP} & 84.84$_{0.28}$  & 83.25$_{0.51}$        & \underline{39.84}$_{5.01}$    & \underline{63.32}$_{1.72}$    & 78.36$_{1.06}$        & 84.53$_{0.48}$        & 56.08$_{3.26}$        & 68.14$_{1.39}$         & \underline{69.79}$_{1.02}$   \\
{\cellcolor{na}\oursAPL} & 84.91$_{1.41}$         & \textbf{83.56}$_{0.59}$       & 39.81$_{2.55}$        & \textbf{64.12}$_{2.45}$       & \underline{79.28}$_{0.63}$    & \textbf{85.26}$_{0.70}$       & 54.07$_{3.74}$         & \textbf{68.87}$_{0.41}$       & \textbf{69.98}$_{0.42}$      \\

\midrule
\multicolumn{10}{l}{[$K=1000$]   Dev Performance}\\
Fine-tuning & 87.70$_{0.89}$     & 84.73$_{0.61}$        & 42.61$_{2.62}$        & 64.90$_{2.01}$        & 78.86$_{2.00}$        & 81.31$_{1.39}$        & 63.74$_{1.59}$        & \textbf{71.99}$_{1.59}$        & 71.98$_{0.59}$       \\
BitFit & 86.30$_{1.36}$  & 83.63$_{0.18}$        & 20.45$_{16.56}$       & 64.24$_{1.55}$        & 76.76$_{0.84}$        & 66.65$_{0.87}$        & 53.22$_{1.73}$        & 68.95$_{2.32}$        & 65.02$_{2.12}$        \\
Adapter & 87.06$_{1.44}$         & 84.79$_{0.42}$        & 43.48$_{1.46}$        & 65.62$_{0.93}$        & 79.88$_{1.26}$        & 80.88$_{1.89}$        & 59.56$_{2.46}$        & 70.52$_{1.48}$        & 71.47$_{0.33}$        \\
Prefix-tuning & \underline{87.86}$_{1.23}$       & 83.48$_{1.15}$        & 44.04$_{2.74}$        & \textbf{68.08}$_{0.81}$       & 79.60$_{1.61}$        & 75.47$_{2.92}$        & \underline{65.48}$_{0.48}$     & 68.94$_{0.93}$        & 71.62$_{0.54}$       \\
LoRA & 87.50$_{1.01}$    & 85.09$_{1.02}$        & \textbf{47.11}$_{3.02}$       & \underline{67.20}$_{0.78}$    & \textbf{80.86}$_{1.88}$       & 76.33$_{1.28}$        & 62.86$_{1.53}$        & \underline{71.48}$_{1.45}$     & 72.30$_{0.52}$       \\
{\cellcolor{na}\oursAP} & 87.32$_{1.73}$  & \underline{85.52}$_{0.63}$    & 45.48$_{3.52}$        & 66.60$_{0.99}$        & \underline{80.70}$_{1.59}$    & \underline{82.96}$_{1.47}$    & \textbf{65.56}$_{2.09}$        & 70.58$_{1.44}$        & \underline{73.09}$_{0.46}$   \\
{\cellcolor{na}\oursAPL} & \textbf{88.02}$_{1.28}$        & \textbf{86.05}$_{0.73}$       & \underline{45.70}$_{2.47}$    & 66.86$_{1.32}$        & 80.50$_{1.76}$        & \textbf{83.09}$_{1.55}$       & 64.60$_{0.72}$         & 70.64$_{0.77}$        & \textbf{73.18}$_{0.27}$      \\

\cmidrule[0.01em]{2-10}
\multicolumn{10}{l}{[$K=1000$]  Test Performance}\\
Fine-tuning & \underline{86.54}$_{1.01}$         & 84.87$_{0.64}$        & 43.26$_{2.60}$        & 62.31$_{2.10}$        & 79.03$_{1.11}$        & 86.39$_{0.34}$        & 61.95$_{1.20}$        & \textbf{71.09}$_{0.77}$        & 71.93$_{0.37}$       \\
BitFit & 83.99$_{0.39}$  & 83.95$_{0.81}$        & 22.44$_{17.10}$       & 62.89$_{1.40}$        & 77.43$_{0.53}$        & 79.04$_{0.61}$        & 52.87$_{0.72}$        & 69.50$_{0.16}$        & 66.51$_{2.22}$        \\
Adapter & 85.60$_{0.63}$         & 84.49$_{0.60}$        & 42.33$_{1.98}$        & 61.81$_{1.57}$        & 79.68$_{0.23}$        & 85.52$_{0.29}$        & 57.86$_{2.44}$        & 70.32$_{0.71}$        & 70.95$_{0.55}$        \\
Prefix-tuning & 85.09$_{0.99}$   & 83.66$_{1.82}$        & 44.07$_{2.90}$        & \textbf{66.71}$_{2.72}$       & 80.34$_{0.70}$        & 82.38$_{1.25}$        & \textbf{63.59}$_{1.12}$       & 68.58$_{0.35}$         & 71.81$_{0.52}$       \\
LoRA & 86.26$_{1.22}$    & \underline{86.04}$_{0.99}$    & \textbf{45.50}$_{1.11}$       & \underline{65.63}$_{2.11}$    & \underline{81.00}$_{0.98}$    & 81.56$_{1.97}$        & 61.32$_{1.65}$        & 70.89$_{0.81}$         & 72.28$_{0.69}$       \\
\cellcolor{na}\oursAP & 86.17$_{0.37}$  & 85.86$_{1.05}$        & 44.33$_{3.55}$        & 64.91$_{1.92}$        & 80.65$_{0.57}$        & \underline{86.82}$_{0.23}$    & 62.17$_{0.99}$        & 69.95$_{0.90}$ & \underline{72.61}$_{0.53}$   \\
\cellcolor{na}\oursAPL & \textbf{87.06}$_{0.81}$        & \textbf{86.65}$_{1.10}$       & \underline{45.44}$_{1.97}$    & 65.49$_{1.92}$        & \textbf{81.22}$_{0.51}$       & \textbf{87.10}$_{0.21}$       & \underline{62.49}$_{0.94}$     & \underline{70.99}$_{0.95}$    & \textbf{73.31}$_{0.52}$      \\

\bottomrule
\end{tabular}}
\upv
\caption{Results on the GLUE benchmark with $K=\{100, 500, 1000\}$ training samples. The evaluation metrics are Matthew's Correlation for CoLA, F1 for MRPC and QQP, Spearman's correlation for STS-B, and accuracy for the rest. For MNLI, we evaluate on the matched dataset. We report average performance on five random seeds with standard deviation as the subscript. \textbf{Best} and \underline{2nd best} methods under each setup are bold and underlined.} 
\label{table_glue_lowres_detail}
\downv
\end{table*}

\end{document}